\documentclass[10pt,a4paper,conference]{IEEEtran}
\usepackage{hyperref}
\usepackage{url}
\usepackage{times}
\usepackage{epsfig}
\usepackage{color}
\usepackage{amsmath}
\usepackage{amssymb}
\usepackage[]{algorithm2e}
\usepackage{caption}
\usepackage{subcaption}

\usepackage{tablefootnote}
\usepackage{cleveref}
\usepackage{footmisc}

\renewcommand{\baselinestretch}{0.99}

\ifCLASSOPTIONcompsoc
  \usepackage[nocompress]{cite}
\else
  \usepackage{cite}
\fi
\ifCLASSINFOpdf
\usepackage{graphicx}
\else
  \usepackage{graphicx}
\fi
\title{Integrating Deep Features for Material Recognition}

\author{\IEEEauthorblockN{Yan Zhang, Mete Ozay, Xing Liu and Takayuki Okatani}
\IEEEauthorblockA{Graduate School of Information Sciences, 
Tohoku University\\
Aramaki Aza Aoba, Aoba-ku, 980-8579 Sendai, Japan\\ \texttt{\{zhang, mozay, ryu, okatani\}@vision.is.tohoku.ac.jp}}
}

\begin{document}

\maketitle

\begin{abstract}
This paper considers the problem of material recognition. Motivated by an observation that there is close interconnection between material recognition and object recognition, we study how to select and integrate multiple features obtained by different models of Convolutional Neural Networks (CNNs) trained in a transfer learning setting. To be specific, we first compute activations of features using representations on images to select a set of samples which are \textit{best} represented by the features. Then, we measure uncertainty of the features by computing entropy of class distributions for each sample set. Finally, we compute contribution of each feature to representation of classes for feature selection and integration. Experimental results show that the proposed method achieves state-of-the-art performance on two benchmark datasets for material recognition. Additionally, we introduce a new material dataset, named EFMD, which extends Flickr Material Database (FMD). By the employment of the EFMD for transfer learning, we achieve $\mathbf{84.0\%\pm1.8\%}$ accuracy on the FMD dataset, which is close to reported human performance $\mathbf{84.9\%}$.



\end{abstract}

\section{Introduction}

In this work, we consider the problem of material recognition, which is to identify material categories such as \textit{glass} or \textit{fabric} of an object (i.e., the material from which the object is made) from its single RGB image. We are particularly interested in interconnection between material recognition and object recognition, by utilizing which we aim to perform material recognition accurately. To be specific, we want to develop a method that can efficiently transfer feature representations learned on different tasks/datasets including object recognition to a target task of material recognition. Toward this end, using convolutional neural networks (CNNs) \cite{alexnet,vgg}, we study how to select and integrate multiple features obtained by different models of CNNs trained on different tasks/datasets.

\begin{figure}[t] \footnotesize
\centering
\begin{subfigure}[b]{.452\textwidth}
   \includegraphics[width=\textwidth]{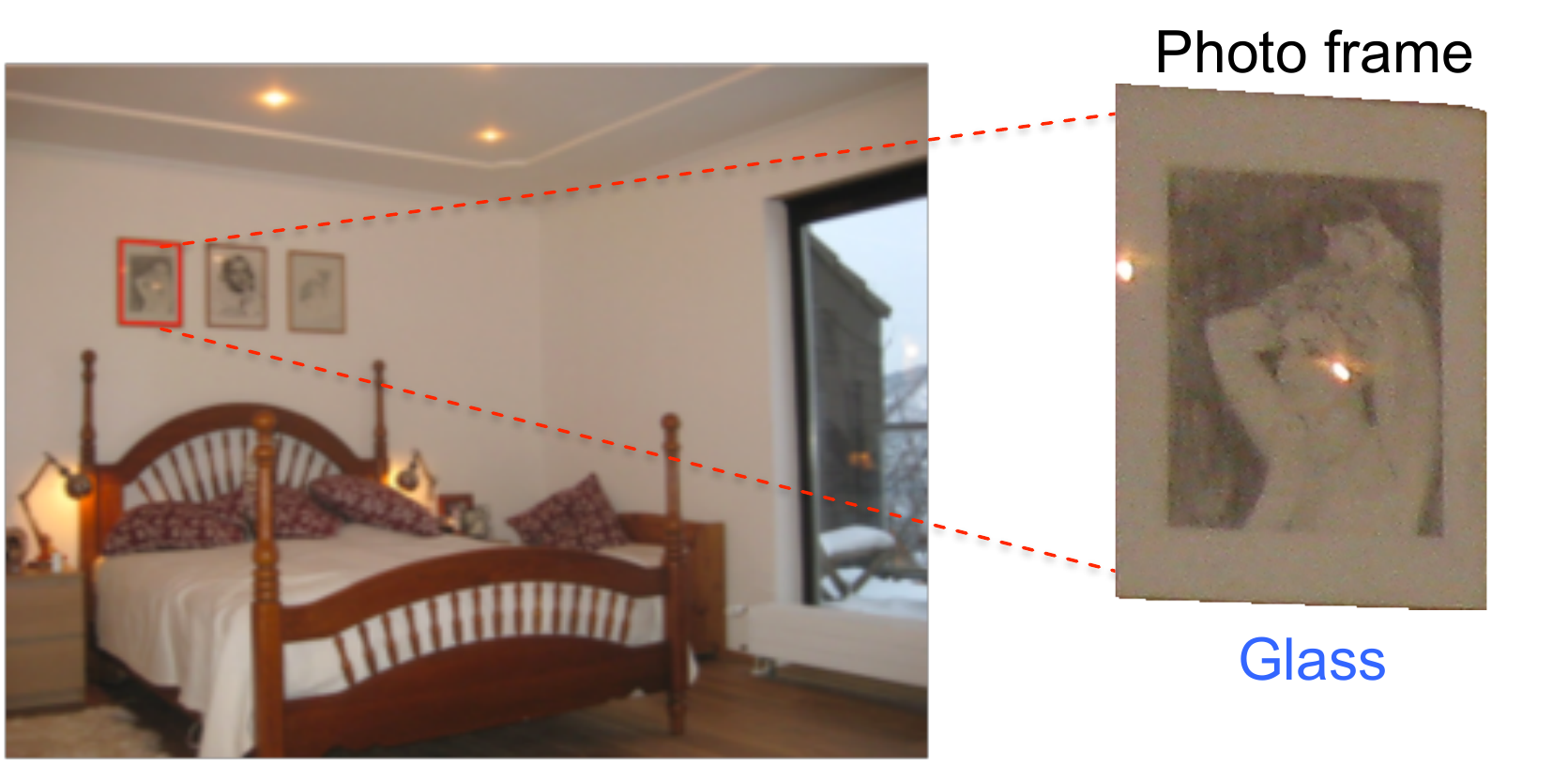}
   \vspace*{-4mm}
   \caption{}
   \label{fig:scene1}
\end{subfigure}

\medskip
\begin{subfigure}[b]{.45\textwidth}
   \includegraphics[width=\textwidth]{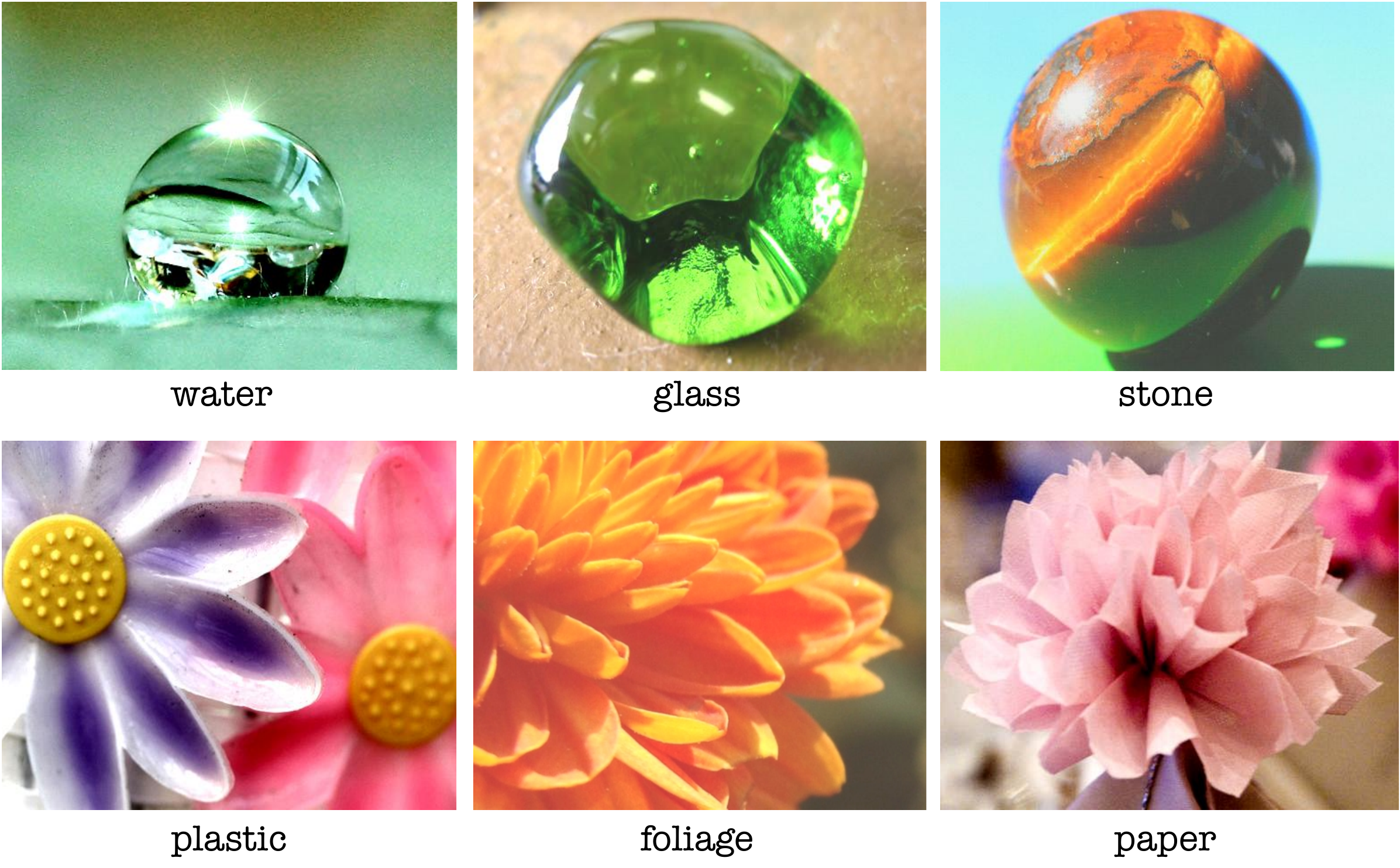}
   \vspace*{-4.5mm}
   \caption{}
   \label{fig:fmd1}
\end{subfigure}%
\caption{(a) An example of integrating knowledge of object and material representations for material recognition. We can first recognize the target image (the image patch within the red rectangle) as a photo frame (object), and then we infer the photo frame as \textit{glass} category (material). (b) As shown in the first row, \textit{circle} shape appears in three different material categories. The second row shows that features representing \textit{flowers} are not merely specific for \textit{foliage}.  }
\end{figure}

A few studies of
psychophysics \cite{Sh1,Nakauchi} imply that material perception in
human vision is interconnected with perception of object category. An observation is that human can perceive some material properties and material category of objects only after correct recognition of the object categories. An example is shown in Fig.~\ref{fig:scene1}. This dependency can be reversed;  object
category of some objects can be correctly recognized only after accurate perception of 
material category of the objects. In this work, we conjecture that there exist
mutual dependencies between perception of object and material. This
could be further generalized to wider contexts such as perception of
scene and texture, although we focus on the relationship between 
object and material categories in this work.

However, it is challenging to model such dependencies and utilize them to accurately perform the task (material recognition in our case). For example, it is possible that an image which belongs to an object category \textit{flower} will be classified as one of the material categories, \textit{plastic}, \textit{paper}, and \textit{foliage} (see Fig.~\ref{fig:fmd1}). In addition to such category-level dependency, we consider representation-level dependency, which will be much more complicated. It is evident that not every learned representation for object recognition is beneficial to material recognition,  
due to divergent appearance that a material category may exhibit. Hence, it is important to select \textit{useful} object and material features for the material recognition task. 

In this paper, we propose a feature selection method to select and combine deep features learned in a transfer learning setting. The contributions of the paper are summarized as follows:
\begin{itemize}
\item We propose a method for material recognition by selecting and integrating multiple features of different CNN models. They are pre-trained on different datasets/tasks and, if possible and necessary, further fine-tuned on the target dataset/task in advance.

\item We introduce an extended version of the benchmark material dataset (namely, FMD \cite{fmdcite}), called EFMD. EFMD is ten times larger than the FMD dataset while the images of EFMD are selected to provide surface properties that are similar to the images of FMD. By the employment of EFMD for transfer learning, we achieve $84.0\%\pm1.8\%$ accuracy on FMD, which is close to human performance ($84.9\%$)  \cite{Sh2}.
\end{itemize}

The rest of this paper is organized as follows. We summarize some related work in Section \ref{relatedwork}. The proposed method is introduced in Section \ref{method}. Section \ref{experiment} provides experiments conducted on several benchmark datasets. Section \ref{conclusion} concludes this study.

\section{Related Work} \label{relatedwork}
Surface properties of materials, and the relationship between perception of material and object categories are analyzed in \cite{Sh2,Sh1}. They first proposed a well-designed benchmark dataset called FMD. Then, they designed descriptors to extract hand-crafted features for representation of various surface properties such as color, texture and shape for material recognition in \cite{Sh2}. Moreover, they analyzed the relationship between object and material recognition for accurate and fast perception of materials in \cite{Sh1}. 

In \cite{Cimpoi1}, a filter bank based method was developed using CNNs for texture recognition. The authors achieved  state-of-the-art performance on several benchmark datasets for texture recognition and material recognition, including 82.4\% accuracy on the FMD dataset. In \cite{Schwartz}, a method was proposed to discover local material attributes from crowdsourced perceptual material distances. They show that without relying on object cues (e.g., outlines, shapes), material recognition can still be performed by employing the discovered local material attributes. In this work, we focus on utilizing object cues for material recognition.

Commonly used feature selection methods can be categorized into wrapper methods \cite{fswrapper0,fswrapper1,fswrapper2} and filter methods \cite{fsfilter0,fs3,fs2,fs1,mrmr,fsif1,fsif2}. Wrapper methods are used to select features by training a classifier on a subset of features, and determine the utility of these features according to the accuracy of the classifier. Features are ranked according to particular criteria using filter methods. For instance,  correlation coefficients were used in \cite{fs3}  to measure importance of each individual feature. Mutual information is also a popular criterion used for ranking features \cite{fsif1,fsif2,mrmr}. A max-relevance and min-redundancy criterion was proposed in \cite{mrmr}. However, in this method, they compute mutual information of each pair of features at each iteration to evaluate redundancy of features. Therefore, the computation cost of the algorithm was increased with the number of candidate features. In \cite{vj}, a feature selection method was proposed based on a variant of Adaboost. In this method, each feature is considered as a weak classifier. At each iteration, a weak classifier that performs best is selected, and the weights assigned to samples are updated. An advantage of boosting based feature selection methods is running time for the training phase. In this paper, we propose a filter-based feature selection method in order to  select the most discriminative features efficiently by minimizing an entropy criteria in a boosting scheme.

\section{Feature Selection and Integration for Deep Representations} \label{method}

In this section, we propose a method to analyze and integrate multiple features extracted by different CNNs pretrained on different tasks. 

\subsection{Outline of the method}

Specifically, we first extract features from material images using multiple CNN models trained on different datasets/tasks. Here, by a feature we mean an activation value of a unit (neuron) at a chosen layer of the CNN. If it is possible and necessary, the CNN models may be fine-tuned on the training samples of the target task of material recognition. Similarly to boosting in \cite{vj}, we then consider each individual feature as a weak classifier, although we do not explicitly train a weak classifier for each feature. Instead, we utilize the class entropy as criteria for measuring how discriminative each feature is. As suggested in \cite{top1,top2}, the images maximizing a feature are the most representative samples for utilization of the feature. Thus, we calculate class entropy for each feature using a set of images on which the activation value of the feature is maximized, and then use a weighted sum of the class entropy over the image set. A weight is given to each sample in the training data. Starting with equal weights for all the samples, the proposed method updates the weights and selects features in an iterative way. At each iteration, the most discriminative feature is selected for integration according to the weighted class entropy. Meanwhile, the weights of images on which an integrated feature is activated are penalized according to the class entropy. This procedure enables us to select a set of the most discriminative features for a given length of features.


\subsection{Details of the algorithm}

Suppose that we are given $N$ image datasets ${\mathcal{D}_n=\{(I^{(i)},Y^{(i)})\}_{i=1}^{M_n}}$, $n=1,2,\ldots,N$, where  $I^{(i)} \in \mathcal{X}_n$ is the $i^{th}$ image, $Y^{(i)} \in \mathcal{Y}_n$ is the corresponding class label, and the samples $\{(I^{(i)},Y^{(i)})\}_{i=1}^{M_n}$ are independent and identically distributed (i.i.d.) according to a distribution $\mathcal{P}_n$ on $\mathcal{X}_n \times \mathcal{Y}_n$. 
We then train a CNN using each dataset $\mathcal{D}_n$ for the associated task. The CNNs may have an identical network architecture or  different ones. Choosing a layer (or a combination of layers) of each CNN trained on $\mathcal{D}_n$\footnote{In our experiments, we used VGG-D-16 \cite{vgg} and the fc7 layer.}, we use $\Phi_n$ to denote the mapping from an input image to the activation value of the layer(s) computed by the CNN.
Given a new dataset ${\mathcal{D}_b=\{(I^{(i)},Y^{(i)})\}_{i=1}^{M_b}}$, we express the activation value of an image $I^{(i)} $ of the sample $(I^{(i)},Y^{(i)}) \in \mathcal{D}_b$ as $\mathbf{x}^i_{nb} = \Phi_n(I^{(i)})$. 

In this work, we aim to extract features from a given set of material images $\mathcal{D}_b$ by employing different CNNs independently trained using images of objects and materials. Therefore, we consider only datasets $\mathcal{D}_n$'s consisting of images of objects or materials. For the sake of simplicity of the notation and concreteness, we denote $\mathbf{x}_m \triangleq \Phi_n(I \in \mathcal{D}_b)$ and $\mathbf{x}_o \triangleq \Phi_{n'}(I \in \mathcal{D}_b)$  as  features extracted from an image $I \in \mathcal{D}_b$ using $\Phi_{n}$ learned on a dataset of material images $\mathcal{D}_n$, and using $\Phi_{n'}$ on a dataset of object images $\mathcal{D}_{n'}$, respectively. Then the feature vectors $\mathbf{x}_o$ and $\mathbf{x}_m$ are concatenated to obtain a feature vector  $ \mathbf{x_c} = [\mathbf{x}_o, \mathbf{x}_m]$.

In the proposed method, we model class discriminative features using a classifier that will be employed on concatenated features $\mathbf{x}_{c}$
by improving discriminative properties of features $\mathbf{x}_{m}$ and $\mathbf{x}_{o}$,
for classification of material images belonging to a dataset $\mathcal{D}_b$. For this purpose, we consider each individual feature as a weak classifier as suggested in \cite{vj}. However, we do not explicitly train a weak classifier for each feature. Instead, we first analyze the discriminative property of each individual feature in $\mathbf{x}_{m}$ and $\mathbf{x}_{o}$.

In CNNs implemented using ReLU activation functions ( \cite{alexnet,vgg}), the features that are represented in an individual neuron are visualized by analyzing and selecting the images on which the activation value of the neuron is maximized \cite{top1,top2}. In our method, we also employ this idea to measure contribution of features to discrimination of classes. Suppose that we are given a set of concatenated features $\mathcal{C} = \{(\mathbf{x}^i_c,Y^i) \}_{i=1}^{M_b}$ extracted from $\mathcal{D}_b$. The $j^{th}$ individual feature $x_{c,j}$ of $\mathbf{x}^i_c$ is investigated by searching for top $K$ samples that have maximal positive feature values on $x_{c,j}$:
\begin{equation}
\begin{split}
 \mathcal{T}_j=\{(\mathbf{x}^k_c,Y^k)\} _{k=1}^K \\
 s.t.\quad x^k_{c,j}>x^i_{c,j}, \quad x^k_{c,j}>0\\
 \forall(\mathbf{x}^k_c,Y^k) \in\mathcal{T}_j \\
\forall(\mathbf{x}^i_c,Y^i) \in \mathcal{C} \setminus \mathcal{T}_j .
\end{split}
\label{eq:top}
\end{equation}


Note that, the samples belonging to $\mathcal{T}_j$ can reveal some properties of the feature $x_{c,j}$.
Then, we compute the discriminative property of $x_{c,j}$ by computing the class entropy in the set $\mathcal{T}_j$ as
\begin{equation}
H_j(Y)=-\sum\limits_{(\mathbf{x}_c,Y) \in \mathcal{T}_j } p(Y)\log {p(Y)}.
\label{eq:th}
\end{equation}



In the next part of the proposed method, we select most discriminative features for discrimination of the whole dataset. Given a set of individual features $\mathcal{F}=\{x_{c,j} \in \mathbf{x_c}\}_{j=1}^{|\mathbf{x_c}|}$, we initialize a set of features to be integrated $\mathcal{S} = \emptyset$, and assign equal weights $w_i=1$ to each sample $(\mathbf{x_c}^{i},Y^{i})\in \mathcal{C}$. Next, we select a feature that has minimum weighted class entropy as
\begin{equation}
    x_{c,\sigma}=\operatornamewithlimits{arg\,min}_{x_{c,j} \in \mathcal{F}} \sum\limits_{(\mathbf{x_c}^{k},Y^{k}) \in \mathcal{T}_j} w_k*H_j(Y),
\label{eq:select}
\end{equation}
and update the sets $
    \mathcal{S}=\mathcal{S} \cup \{x_{c,\sigma}\}$, $
    \mathcal{F}=\mathcal{F} \setminus \{x_{c,\sigma}\}
$. Then, we penalize the top $K$ samples $(\mathbf{x_c}^{k},Y^{k}) \in \mathcal{T}_j$ of the integrated feature using
\begin{equation}
w_k=w_k*(1+H_j(Y)^{-1}).
\label{eq:punish}
\end{equation}
A pseudo-code of our algorithm is shown in Algorithm 1.


\RestyleAlgo{boxed}\LinesNumbered
\begin{algorithm}		
\Indp \SetKwInOut{Input}{Input}
 \SetKwInOut{Output}{Output}
 \SetKwInOut{Initialization}{Initialization}
\caption{Integration of features extracted using deep representations of CNNs.}
\Input{ 
	
	- $\{ \Phi_n \}_{n=1}^N$:
	A set of representations each of which is learned using a CNN obtained on a dataset $\mathcal{D}_n$. 
	\newline
	- $\mathcal{D}_b=\{(I^{(i)},Y^{(i)})\}_{i=1}^{M_b}$: A dataset of images  that will be used for inference of representations. 
	\newline
- $T$: The number of integrated features.
\newline
- $K$: The number of samples that have maximal activation values on each feature.}
\Output{
	
- $\mathcal{S}$: A set of integrated features.}
\BlankLine
\Initialization {
	
	- For each $(I,Y)\in \mathcal{D}_b$, extract features using a representation $\Phi_n(\cdot)$ such that $\mathbf{x}_n=\Phi_n(I)$, $\forall n=1,2,\ldots,N$.
\newline
- Concatenate $\mathbf{x}_n$, $\forall n$ and construct  $\mathbf{x}_c = [\mathbf{x}_n]_{n=1}^N$.
\newline
- Define $\mathcal{C} = \{(\mathbf{x}^i_c,Y^i) \}_{i=1}^{M_b}$, and set equal weight $w_i=1$ to each sample in $\mathcal{C}$. 

- For each individual feature of $\mathbf{x_c}$, define the sets  $\mathcal{F}=\{x_{c,j} \in \mathbf{x_c}\}_{j=1}^{|\mathbf{x_c}|}$, and 
 $\mathcal{S}=\emptyset$.}
\For{$j \leftarrow 1$ \KwTo $|\mathcal{F}|$}{
     Construct $K$ samples that have maximal feature values on $x_{c,j}$ using \eqref{eq:top};\\
    Compute class entropy on the set of top $K$ samples using \eqref{eq:th};
}

\For{$t\leftarrow 1$ \KwTo $T$}{
    Normalize the sample weights: $w_i^{t+1}= \frac{w_i^t}{\sum\limits_{j=1}^{M_b} w_j^t}, \forall i$;\\
    Select the feature that minimizes the weighted class entropy using \eqref{eq:select};\\
    Penalize the samples using \eqref{eq:punish};\\
}
\end{algorithm}







\section{Experimental Analysis} \label{experiment}

\subsection{Datasets}
\begin{table}[t]\footnotesize
\caption{Datasets used in experiments.}
\begin{center}
\begin{tabular}{|l|l|l|l|l|l|}
\hline
Datasets & \# categ. & \# samples & \# train &\#  test & \# splits\\
\hline\hline
FMD   & 10 & 1,000  & 500   & 500   & 10\\
EFMD         & 10 & 10,000 & 5,000   & 5,000  & 10\\
MINC         & 23 & 3 Million & Training & Val./Test & -\\
\hline
\end{tabular}
\end{center}
\label{table:1}
\end{table}

%
In the experiments, we used three benchmark datasets (see Table \ref{table:1}). FMD dataset \cite{fmdcite} consists of 10 material categories each of which contains 100 images. In the FMD, the samples were selected manually from Flickr covering different illumination conditions, compositions, colors, and texture. MINC is a large scale material recognition dataset \cite{MINC}. It contains 3 million images belonging to 23 categories. 

In addition, we introduce a new dataset called EFMD which is an extended version of the FMD. EFMD consists of the same categories that are used in FMD, each of which contains 1,000 images. Thus, the size of EFMD is 10 times larger than the size of FMD. While constructing EFMD, we try to make it as \textit{similar} to FMD as possible in the context of visual perception and recognition. Specifically, we first pick 100,000 images from Flickr by text searching. Then we ask Amazon Mechanical Turkers to choose the images that are \textit{similar} to FMD images. In each Human Intelligence Task, we present 10 candidate images to a Turker along with three good examples (FMD images) and six bad examples, and ask the Turker to select good ones. Publishing each Human Intelligence Task for three Turkers, we only select images that are selected by all the Turkers. We then manually crop these images to adjust the scale of an object appearing in the images. Note that the images belonging to FMD will not be selected and merged into EFMD to make sure that there is no overlapping between them. 

Each of FMD and EFMD datasets is randomly split into two subsets of equal size; one is used for training and the other is used for testing. This scenario is performed 10 times and the average classification accuracy was reported. 
For MINC dataset, we directly use the originally provided train, validation and test sets.  The details of each dataset and the corresponding experimental setup are shown in Table \ref{table:1}.  

\subsection{Details of experimental setups}

In our experiments, we consider four different tasks, which we will refer to as FMD, FMD-2, EFMD, and MINC(-val/test), respectively. In each of them, we construct a material representation $\Phi_n$ and an object representation $\Phi_n'$ in {\em different} ways described below.  

In all the tasks, we start with two CNNs pre-trained on MINC and ILSVRC2012 \cite{ILSVRC15}, for which we use publicly available pre-trained Caffe \cite{jia2014caffe} models of VGG-D consisting of 16 layers \cite{vgg}. 
We then fine-tune these CNNs for the tasks FMD, FMD-2, and EFMD using different samples as shown in Table \ref{tbl:tasks}. The fine-tuning is performed in such a way that the weights of Conv1-Conv4 layers are fixed, and those of the higher layers are updated. We have found that this selection of fixed/updating layers performs the best for FMD and EFMD. We do not conduct fine-tuning on MINC dataset. This is because MINC has a large number of samples that enables to train a CNN from scratch. Although it is possible to fine-tune the CNN pre-trained on ILSVRC2012 using MINC dataset, it will be no longer a 'fine-tuning' because of the size of the dataset, and thus we use the pre-trained model as is. 

\begin{table}[t]
\caption{Two CNNs pretrained on MINC(M) and ILSVRC2012(O) are fine-tuned for each task using the datasets shown below, from which $\Phi_n$ and $\Phi_{n'}$ are obtained, respectively.}
\label{tbl:tasks}
\centering
\begin{tabular}{|c|c|c|} \hline
Task    & Samples for fine-tuning & \# of samples\\ \hline \hline
FMD     & FMD(train) & 500 \\
FMD-2 & FMD(train) + EFMD(all) & 10,500 \\
EFMD    & EFMD(train) & 5,000 \\
MINC-val/test & None & N/A\\ \hline
\end{tabular}
\end{table}

In the inference phase of each task, we extract material features $\mathbf{x}_m \triangleq \Phi_n(I \in \mathcal{D}_b)$ and object features $\mathbf{x}_o \triangleq \Phi_{n'}(I \in \mathcal{D}_b)$  using the representations learned at the fc7 layer (e.g. using fully connected networks at the $7^{th}$ layer) of the two CNNs, respectively. The concatenated object and material features $ \mathbf{x_c} = [\mathbf{x}_o, \mathbf{x}_m]$ are used to obtain a set of  integrated features $\mathbf{x_s}$. We set $K$ to 10\% of the size of training images, and $T$ to  3,000 throughout the experiments. The integrated features are fed into a support vector machines (SVM) with radial basis function (RBF) kernels \cite{svm} for training and testing of classifiers. 

\subsection{Performance analysis}

\begin{table}[t]
\caption{Performance (accuracy) comparison for different tasks. \textit{M}: Material features learned using MINC. \textit{O}: Object features learned using ILSVRC2012. \textit{MO}: Concatenated material and object features ($\mathbf{x}_c \in \mathcal{F}$). \textit{SMO}: Features integrated using the proposed method ($\mathbf{x}_c \in \mathcal{S}$). 
}
\centering
\begin{tabular}{|c|c|c|c|c|c|c|}
\hline
Task & \textit{M} (\%) & \textit{O} (\%) & \textit{MO} (\%)& \textit{SMO} (\%)\\ 
\hline\hline
FMD & $80.4\pm1.9$  & $79.6\pm2.1$ & $79.1\pm2.5$ &$\mathbf{82.3\pm1.7}$ \\
FMD-2 & $82.5\pm2.0$ & $82.9\pm1.6$  & $83.9\pm1.8$ &$\mathbf{84.0\pm1.8}$ \\ 
EFMD & $88.7\pm0.2$ & $88.8\pm0.3$ &$\mathbf{89.7\pm0.13}$ & $\!\!\mathbf{89.7\pm0.16}\!\!\!\!$ \\
MINC-val &82.45 \cite{MINC}& 68.17 &83.48  & $\mathbf{83.93}$ \\
MINC-test &82.19 \cite{MINC} & 68.04 &83.12 & $\mathbf{83.60}$ \\

\hline
\end{tabular}
\label{table:perf}
\end{table}

The results are shown in Table \ref{table:perf};  i) \textit{O}, ii) \textit{M}, iii) \textit{MO} and iv) \textit{SMO} columns denote the results for the case where only i) object features ($\mathbf{x}_o$), ii) material features ($\mathbf{x}_m$), iii) concatenated features ($\mathbf{x_c} = [\mathbf{x}_o, \mathbf{x}_m]$), and iv) the features integrated by proposed method ($\mathbf{x_s}$), are used, respectively. Several observations can be made from the results. 

Firstly, the concatenated features (\textit{MO}) provide better performance than the individual material (\textit{M}) and object (\textit{O}) features for all the tasks, indicating that object features contribute to the material classification. Next, features integrated by our method (\textit{SMO}) further boost the performance obtained using the concatenated features for the task FMD and MINC-val/test, indicating the effectiveness of the proposed method. However, there is practically no difference in accuracy between \textit{MO} and \textit{SMO} for the task FMD-2 and EFMD, which we will analyze later. 

It should also be noted that $82.3\pm 1.7\%$ for FMD and $84.0 \pm 1.8\%$ for FMD-2 are better than the performances of the state-of-the-art methods reported in the literature. A method proposed in \cite{deepbankijcv} achieves $82.4\pm 1.5\%$ on FMD, where they combine fc7 features and Fisher Vectors (FV) features pooled from the conv5\_3 layer of VGG-D-19. However, we should note that their numbers are obtained in different conditions and thus we cannot make direct comparisons. Differences are that they do not use MINC or EFMD dataset, whereas they use a multi-scale approach, i.e., images with different scales are used to compute FV features. It is nonetheless important to note that $84.0 \pm 1.8\%$ is the best performance for FMD and is close to human vision (84.9\%) \cite{Sh2}. Additionally, $83.93\%/83.60\%$ on MINC val/test also outperform the previous ones \cite{MINC}. In their work, $83.83\%/83.4\%$ are obtained by GoogLeNet and $82.45\%/82.19\%$ are achieved by VGG-D-16. Note that their CNN models are pre-trained on ILSVRC2012\cite{ILSVRC15}.

Finally, we can see from the table that the accuracy on FMD-2 is better than FMD. This can be explained by the use of more training samples in FMD-2, where 10,000 EFMD samples in addition to 500 FMD samples are used for training. This also indicates the similarity of EFMD samples to FMD samples, which is consistent with our intention of the creation of EFMD. That said, the accuracy on EFMD, i.e., training on a half of EFMD and test on the rest, is even more higher, indicating that EFMD is easier than FMD and there is some gap between them. 

Now, we analyze the results that the proposed method (\textit{SMO}) does {\em not} improve the concatenated feature (\textit{MO}) for FMD-2 and EFMD. 
In order to analyze \textit{shareability} of class representations among features, 
we compute average entropy $\hat{H}^{\mathcal{F}} = \frac{1}{|\mathcal{F}|} \sum \limits _{i=1} ^{|\mathcal{F}|} H_i(Y)$ and $ \hat{H}^{\mathcal{S}} =  \frac{1}{|\mathcal{S}|} \sum \limits _{i=1} ^{|\mathcal{S}|} H_i(Y)$.
The results given in Table~\ref{tab:div} show that entropy decreases as we employ the EFMD for training CNNs. However, we observe that the difference between entropy values of $\hat{H}^{\mathcal{F}}$ and $\hat{H}^{\mathcal{S}}$ is $\sim0.21$ for FMD and FMD-2, and $\sim0.22$ for EFMD. 

In order to further analyze the shareability of class representations, we computed diversity of decisions of classifiers employed on features ($\mathbf{x}_m$ and $\mathbf{x}_o$) extracted using individual representations of objects ($O$) and materials ($M$). For this purpose, we employed 
five statistical measures \cite{kunc1, kunc2}, 
namely i) inter-rater agreement ($\kappa$) to measure the level of agreement of classifiers while correcting for chance, ii) $Q$ statistics to measure statistical dependency of classifiers, 
iii) Kohavi-Wolpert variance to measure variance of agreement, iv) measurement of disagreement, v) generalized diversity to measure causal statistical diversity of classifiers.

The results are given in Table~\ref{tab:div}. Note that the performance boosts using the proposed method as the diversity of the classifiers employed on the features extracted using individual representations ($O$ and $M$) increases. For instance, the diversity of classifiers employed on the feature sets extracted using FMD dataset is larger than that of the classifiers employed using FMD-2 and EFMD datasets. In addition, the performance difference between classifiers employed on the integrated features ($SMO$) and the concatenated features ($MO$) is $3.2 \%$, $0.1 \%$ and $0 \%$ for FMD, FMD-2 and EFMD, respectively (see Table~\ref{table:perf}). Therefore, we observe that the performance boost obtained using our proposed method increases as diversity of the classifiers employed on individual representations increases.


\begin{table}[t]
  \centering
  \caption{Average entropy values of distributions of  detections of concatenated and integrated features ($MO$ and $SMO$). In addition, we provide a diversity analysis of decisions of classifiers employed on individual feature sets ($O$ and $M$), where $\downarrow$ ($\uparrow$) indicates that the smaller (larger) the measurement, the larger the diversity.}
    \begin{tabular}{|c|c|c|c|}
    \hline
    Diversity Measures & FMD & FMD-2 & EFMD \\
    \hline
    \hline
    $MO$ ($\hat{H}^{\mathcal{F}}$) & 2.40 & 2.29 & 2.21\\
    $SMO$ ($\hat{H}^{\mathcal{S}}$) & 2.19 & 2.08 & 1.99\\
    $\kappa$ ($\downarrow$) & 0.6471 & 0.7103 & 0.7116 \\
    $Q$ Statistics ($\downarrow$) & 0.9860 & 0.9944 & 0.9996 \\
    Kohavi-Wolpert Variance ($\uparrow$) & 0.0070 & 0.0043 & 0.0002 \\
    Disagreement ($\uparrow$)& 0.0279 & 0.0172 & 0.0008 \\
    Generalized Diversity ($\uparrow$)& 0.3383 & 0.2892 & 0.2795 \\
    \hline
    \end{tabular}
  \label{tab:div}
\end{table}%

\subsection{Robustness analysis}

\begin{figure}[t]
\begin{subfigure}[b]{.25\textwidth}
   \includegraphics[scale=0.225]{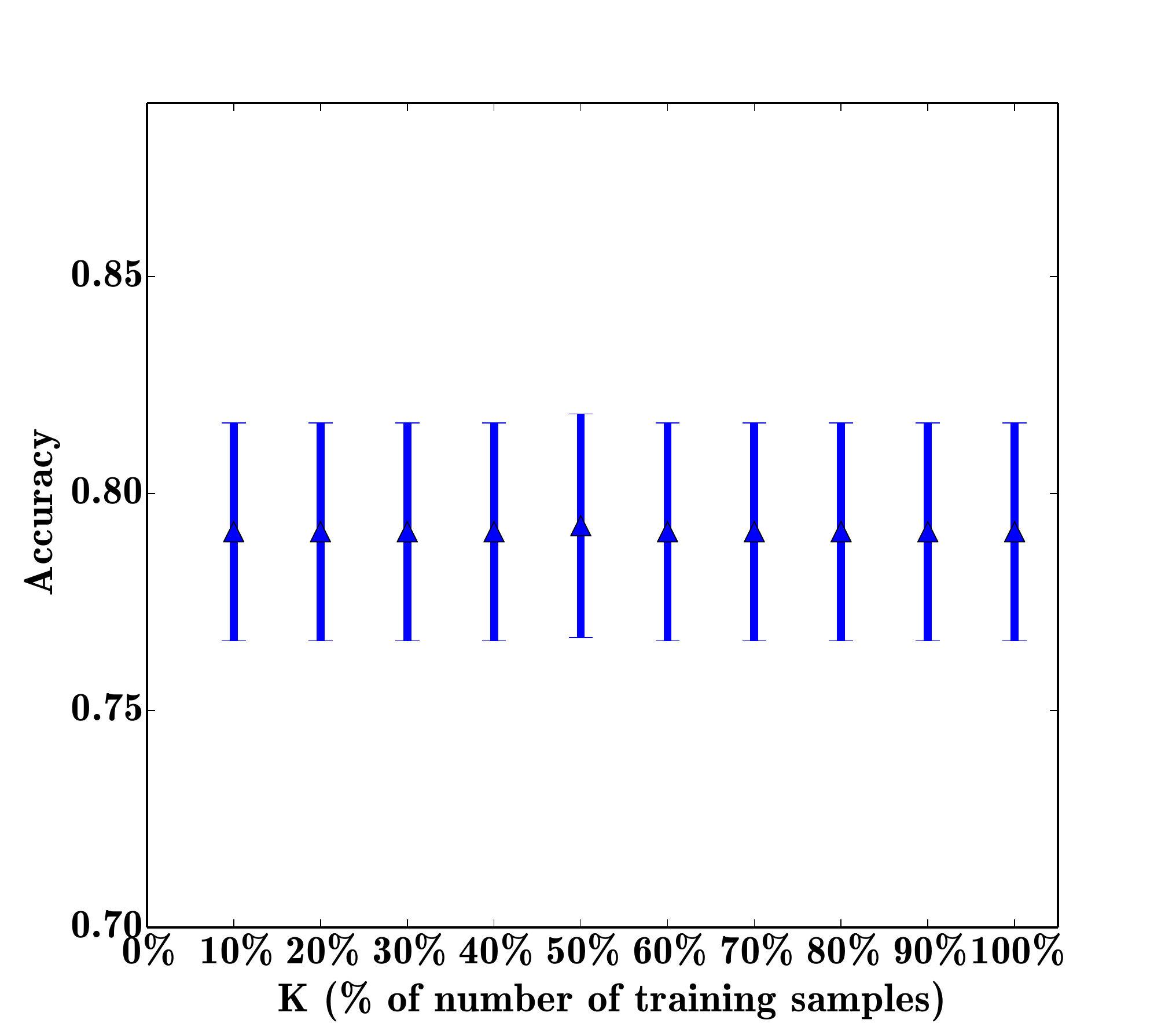}
   \caption{$K$ vs. Accuracy}
   \label{fig:Kacc}
\end{subfigure}%
\begin{subfigure}[b]{.24\textwidth}
   \includegraphics[scale=0.225]{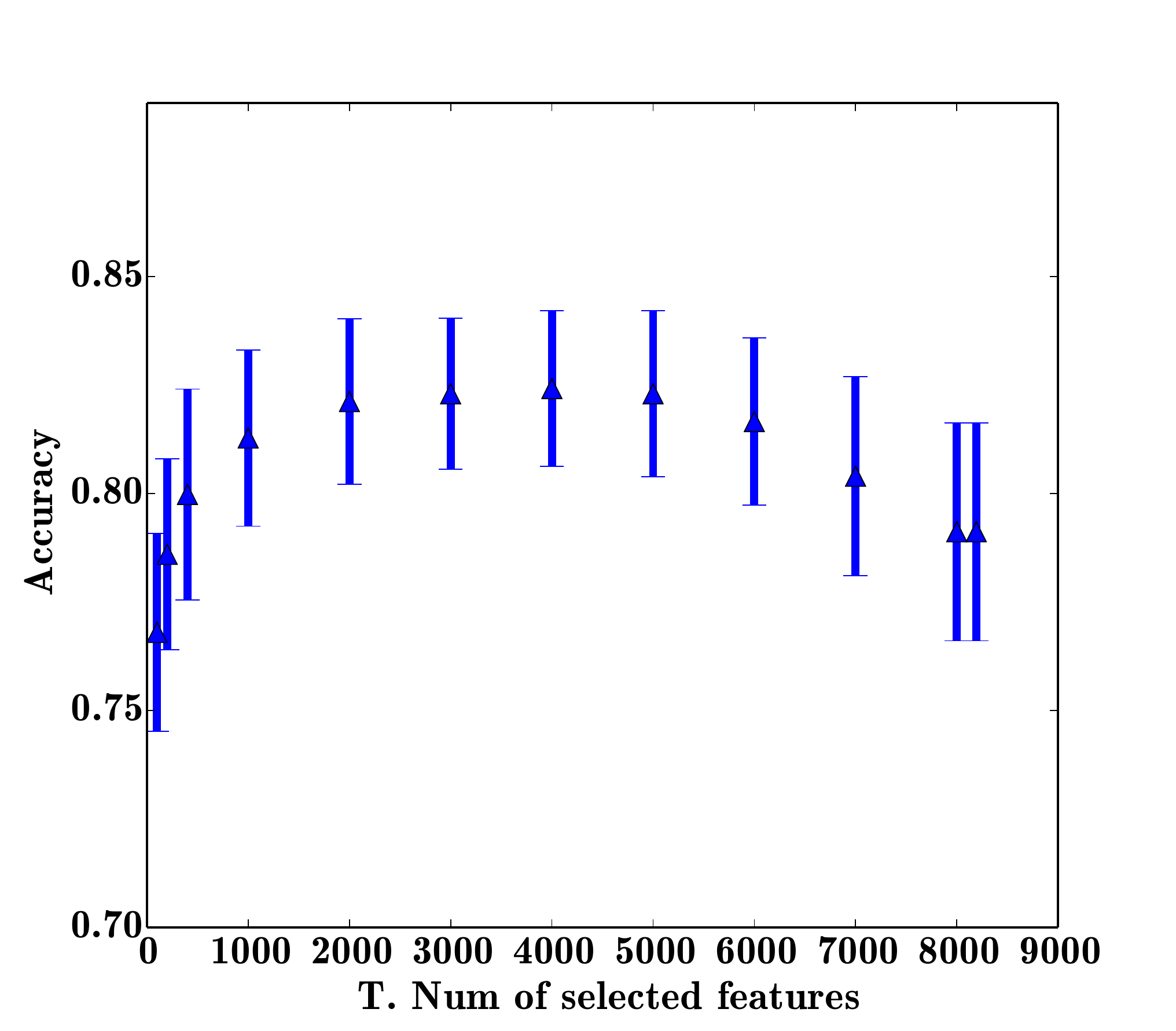}
  \caption{$T$ vs. Accuracy}
   \label{fig:Tacc}
\end{subfigure}%

\caption{Analysis of classification performance for (a) different number of Top-$K$ samples, and (b)  different number ($T$) of integrated features  on FMD. }
\end{figure}

We analyse how the number of Top-$K$ selected samples and the number ($T$) of integrated features affect the classification performance on the FMD dataset; see Fig.~\ref{fig:Kacc} and Fig.~\ref{fig:Tacc}. In Fig.~\ref{fig:Kacc}, the number of integrated features is fixed as 3000. Note that only the samples with positive activation are considered as the top selected samples. For example, if $K$ is selected to cover $100\%$ of the samples, then the ratio of samples selected by the algorithm may be less than $100\%$. This can be observed if most of the activation values of FMD is $0$. In order to analyze the effect of $T$ on the performance, $K$ is fixed as $10\%$. Although selection of $T=3000$ features provides the best performance, less number of integrated features, e.g. 100-400, provides comparable accuracy. 

\begin{figure}
\begin{subfigure}[b]{.25\textwidth}
   \includegraphics[height=38mm]{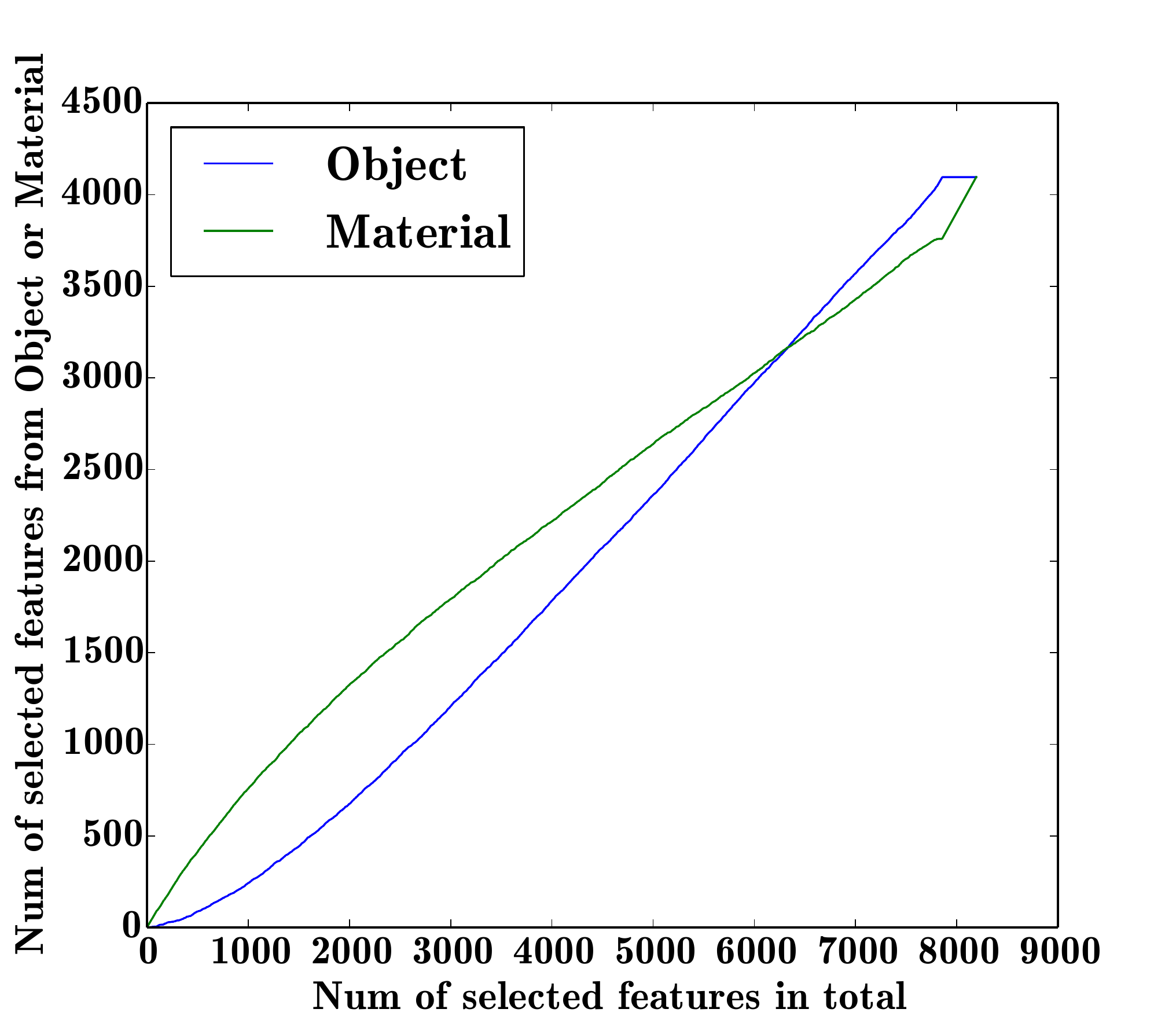}
  \caption{FMD}
   \label{fig:fmdovsm}
\end{subfigure}%
\begin{subfigure}[b]{.25\textwidth}
   \includegraphics[height=38mm]{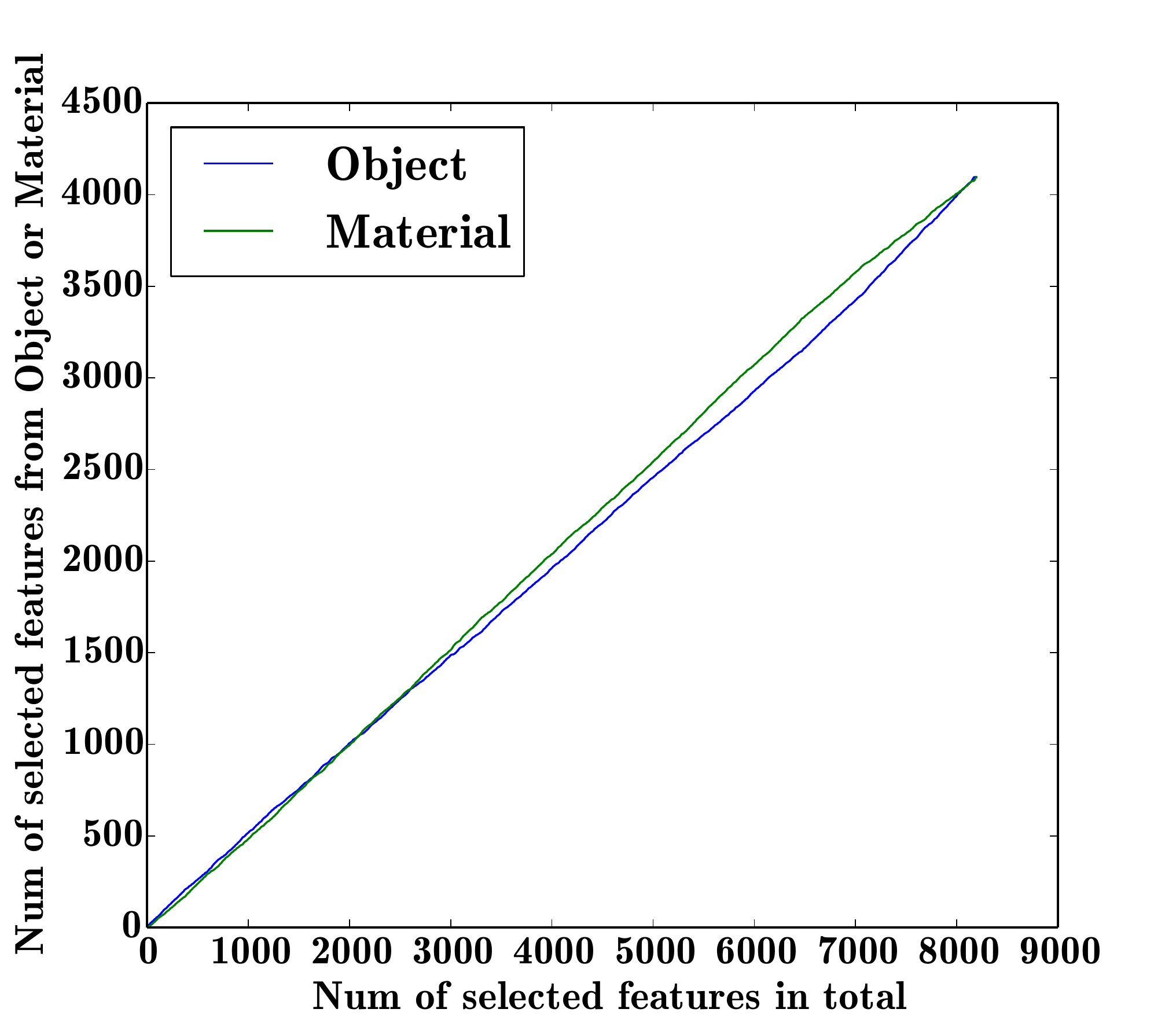}
  \caption{FMD2}
   \label{fig:fmd2ovsm}
\end{subfigure}%
\\
\begin{subfigure}[b]{.25\textwidth}
   \includegraphics[height=38mm]{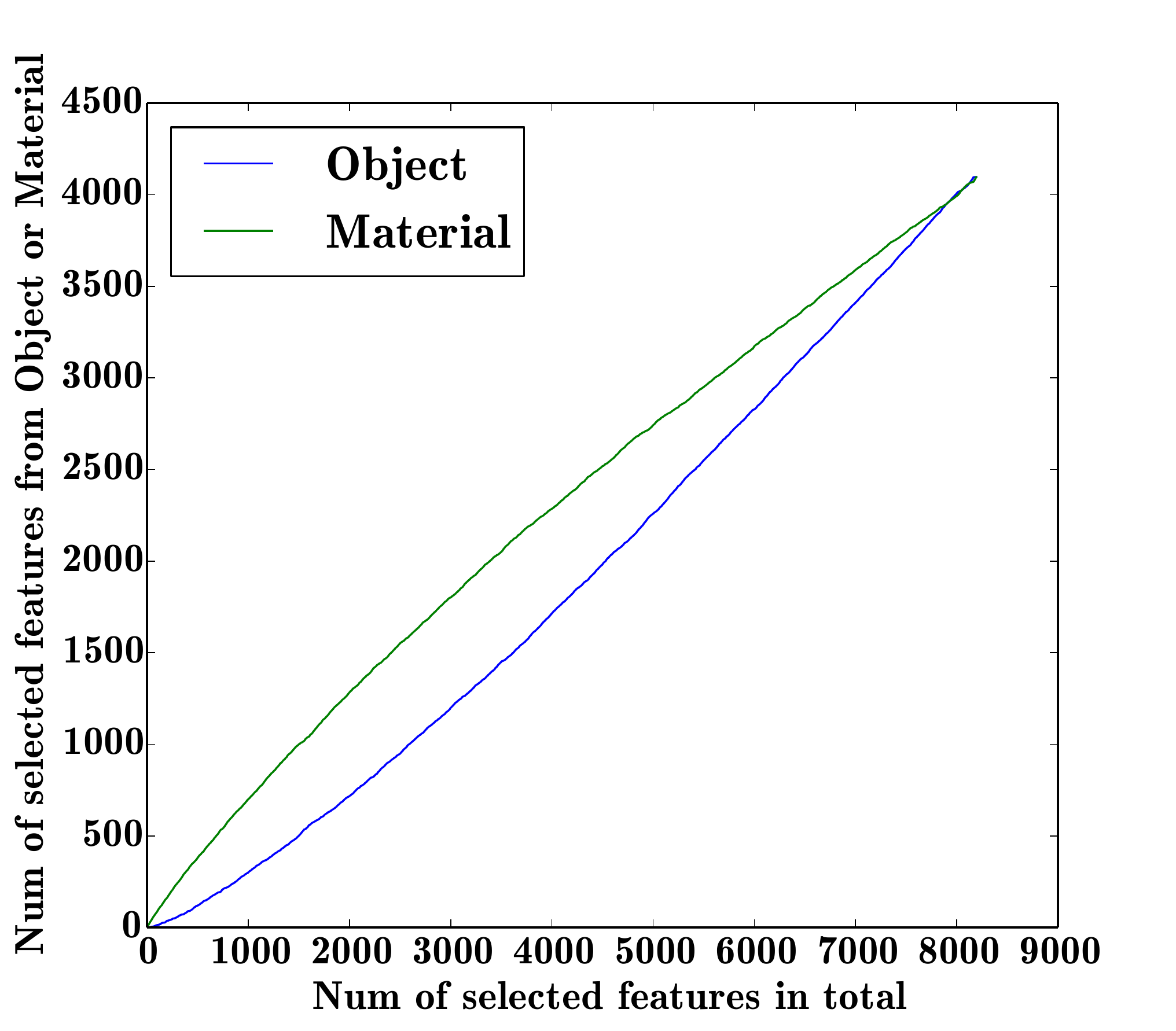}
  \caption{EFMD}
   \label{fig:ufmdovsm}
\end{subfigure}%
\begin{subfigure}[b]{.25\textwidth}
   \includegraphics[height=38mm]{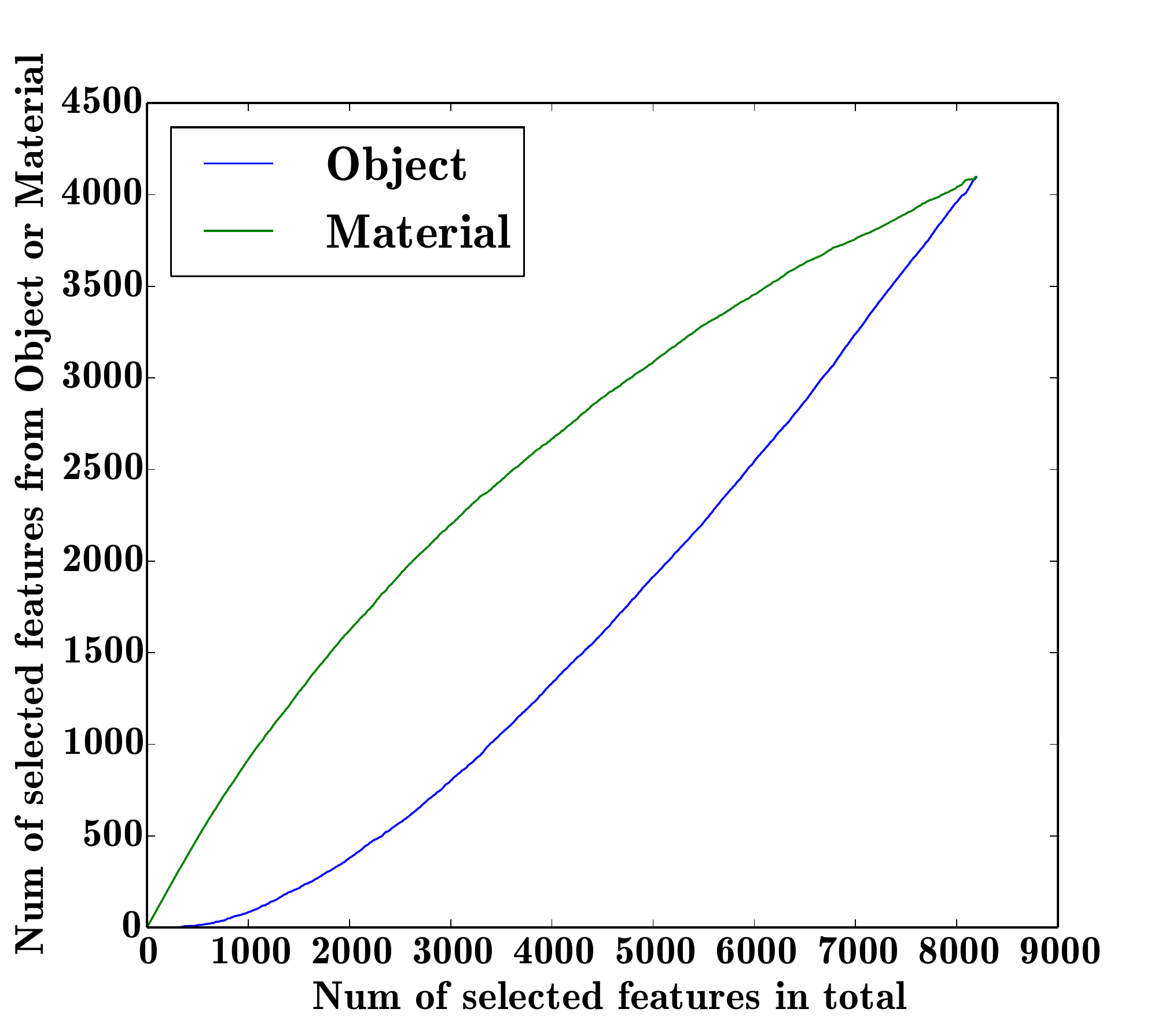}
  \caption{MINC}
   \label{fig:mincovsm}
\end{subfigure}%

\caption{Comparison of number of object and material features belonging to the set of selected features. The number of selected object and material features are shown in (a) FMD, (b) FMD2, (c) EFMD, and (d) MINC.}
\label{fig:ovsm}
\end{figure}

We also analyze the relationship between features that are extracted using representations of materials and objects for feature selection. We leave out one split of FMD, and record the number of integrated material and object features using our feature selection method. A comparative result is shown in Fig.~\ref{fig:ovsm}. As we can see from the figure, there is a gap between the number of selected material  and object features in FMD (Fig.~\ref{fig:fmdovsm}), EFMD (Fig.~\ref{fig:ufmdovsm}) and MINC (Fig.~\ref{fig:mincovsm}). However,  in Fig.~\ref{fig:fmd2ovsm}, we observe that the difference between the number of material and object features is less for FMD2 compared to the other datasets. In FMD2, both object and material features are fully fine-tuned using 10,000 EFMD images. If the fine-tuned features are discriminative for material classification, then the features are equally selected by the proposed method using FMD2. On the other hand, if the features are not \textit{sufficiently} fine-tuned, then material features may be more discriminative than object features for material recognition. For instance,   Table \ref{table:perf} shows the results for the case where material and object features are not fine-tuned for MINC-val. We obtain $82.45\%$ using  individual material features, while we obtain $68.17\%$ using individual object features. Hence, we may obtain larger number of material features than object features as shown in Fig.~\ref{fig:mincovsm}.


\section{Conclusion} \label{conclusion}

In this work, we propose a method to integrate deep features extracted from multiple CNNs trained on images of materials and objects for material recognition. For this purpose, we first employ a feature selection and integration method to analyze and select deep features by measuring their contribution to representation of material categories. Then, the integrated features are used for material recognition using classifiers. In the experimental results, we obtain state-of-the-art performance by employing the features integrated using the proposed method on several  benchmark datasets. In future work, we plan to investigate  theoretical properties of the proposed methods for integration of deep representations using various deep learning algorithms such as autoencoders, to perform other tasks such as scene analysis, image classification and detection.






\bibliography{root}
\bibliographystyle{IEEEtran}

\end{document}